\documentclass[letterpaper, 10pt, conference]{ieeeconf}      

\IEEEoverridecommandlockouts                              

\usepackage{graphics} 
\usepackage{epsfig} 
\usepackage{mathptmx} 
\usepackage{times} 
\usepackage{amsmath}
\usepackage{xcolor}
\usepackage{subcaption}
\usepackage{amssymb}
\usepackage{subcaption}
\usepackage{multicol}
\usepackage{hyperref}
\usepackage[amssymb]{SIunits}
\usepackage{algorithm}
\usepackage{algpseudocode}

\newtheorem{definition}{Definition}
\newtheorem{remark}{Remark}
\newtheorem{assum}{Assumption}

\DeclareSymbolFont{newfont}{OML}{cmm}{m}{it}
\DeclareMathSymbol{\Vatheta}{3}{newfont}{37}

\makeatletter
\newcommand*\titleheader[1]{\gdef\@titleheader{#1}}
\AtBeginDocument{%
  \let\st@red@title\@title
  \def\@title{%
    \bgroup\normalfont\large\centering\@titleheader\par\egroup
    \vskip1.5em\st@red@title}
}
\makeatother

\titleheader{\color{blue}This is a draft version of a paper currently under review. This work has not been published anywhere.}
\title{

Barriers on the EDGE: A scalable CBF architecture over EDGE for safe aerial-ground multi-agent coordination}

\author{Viswa Narayanan Sankaranarayanan$^{*}$, Achilleas Santi Seisa, Akshit Saradagi, Sumeet Satpute, \\and George Nikolakopoulos
\thanks{The authors are with the Robotics and Artificial Intelligence Group of the Department of Computer Science, Electrical and Space Engineering at Lule\aa \ University of Technology, Sweden.}
\thanks{$^*$Corresponding Author's email: {\tt\small vissan@ltu.se}}}
\begin{document}
\maketitle


\begin{abstract}
In this article, we propose a control architecture for the safe, coordinated operation of a multi-agent system with aerial (UAVs) and ground (UGVs) robots in a confined task space. We consider the case where the aerial and ground operations are coupled, enabled by the capability of the aerial robots to land on moving ground robots. The proposed method uses time-varying Control Barrier Functions (CBFs) to impose safety constraints associated with (i) collision avoidance between agents, (ii) landing of UAVs on mobile UGVs, and (iii) task space restriction. Further, this article addresses the challenge induced by the rapid increase in the number of CBF constraints with the increasing number of agents through a hybrid centralized-distributed coordination approach that determines the set of CBF constraints that is relevant for every aerial and ground agent at any given time. A centralized node (Watcher), hosted by an edge computing cluster, activates the relevant constraints, thus reducing the network complexity and the need for high onboard processing on the robots. The CBF constraints are enforced in a distributed manner by individual robots that run a nominal controller and safety filter locally to overcome latency and other network nonidealities.
\end{abstract}
\section{Introduction} \label{sec:intro}
In multi-agent robotics, it is common to see solutions exclusively utilizing either ground or aerial robots. Heterogeneity often translates to the use of heterogeneous robots within the classes of either ground or aerial robots. In this article, we consider a multi-agent system comprising of both aerial and ground robots, specifically, the scenario where aerial and ground operations are not decoupled, but are interacting through the ability of the aerial robots to land on ground robots. Safe deployment of such a coordinated multi-robot system requires the imposition of various safety constraints, both within the ground and aerial layers and between the ground and aerial layers. 
In an aerial-ground multi-agent scenario, three challenges are typically faced: 
(a) reliable imposition of safety, (b) establishing environmental awareness for each agent, and (c) scaling the control architecture for large number of agents \cite{georgeff1988communication, song2024safety}. This article proposes a multi-agent coordination solution that tackles all three challenges, by leveraging i) recent advances in control design approaches that guarantee safety and ii) Edge Computing. 

To guarantee safe operation across the aerial-ground layers, the imposition of several state constraints is crucial. In the context of this article, the constraints arise as i) safety constraints associated with collision avoidance in the aerial and ground layers, ii) constraints associated with safe landing of aerial robots on ground robots, and iii) constraints restricting operation within a pre-specified task space. 
In recent years, Control Barrier Function (CBF) approach has emerged as a popular choice for imposing safety constraints over dynamical systems \cite{CBF_Funda,CBF_Survey}. The CBF approach derives safe control inputs that render a set of safe robot configurations forward invariant, by filtering a nominal goal-reaching controller.  The constraints associated with every robot (ground and aerial) are transformed into safe sets and the corresponding control barrier functions are formulated. Then, for every agent, a time-varying set of CBFs are simultaneously imposed by a CBF filter, in order to render the intersections of the safe sets forward invariant. The CBF-filtered control signal is minimally invasive on the nominal controller and is derived by solving a computationally light quadratic optimization problem (QP) defined over the space of control inputs. 

Popular CBF techniques used in multi-agent coordination involve either centralized \cite{cheng2020safe, lindemann2019control, chen2020guaranteed} or distributed \cite{lyu2023cbf} control approaches, with both 
presenting implementation and theoretical challenges \cite{bertilsson2022centralized}. The main disadvantage of centralized designs is that a major part of the control layer runs offboard, jeopardizing the robot's nominal operation due to networking issues, and creating a single point of failure \cite{gupta2015survey}. Additionally, although centralized control excels in performance, its scalability limitations make it impractical for large-scale multi-agent systems \cite{hu2018centralize}. In contrast, the distributed control approach \cite{tan2021distributed, zhang2023neural} improves the onboard safety of the robots but demands the robots to possess the knowledge of other agents and the environment. While researchers have suggested distributed approaches where agents only interact with their immediate neighbors \cite{wang2017safety}, the scalability of this approach requires awareness of who among the neighboring agents is relevant at any given time, for dynamic establishment and termination of the communication between them.

Gathering knowledge of other agents and the environment constantly can lead to communication overheads, network complexity and heavy onboard processing \cite{gupta2015survey, seisa2022cnmpc}. 
Computational demand can be reduced by determining the set of constraints associated with only the most locally relevant neighbors, which impact an agent. Therefore, we propose a hybrid architecture, which captures the advantages of both strategies. On this line, the Model Predictive Control implementation in \cite{zhan2022data} exploits a hybrid distributed and centralized approach to harness the advantages of both strategies. Inspired by this notion, we propose a hybrid control approach for multi-agent systems that improves the scalability and reduces the need for large onboard computational resources in the proposed CBF-based aerial-ground multi-robot coordination solution. In this setting, Edge Computing, which facilitates large off-board computing resources to be utilized with low latency, has emerged as a practical method to offload a part of the processing \cite{seisa2022edge} from the agents.

\textbf{Contributions:} Given the premise, we present the  contributions of this work next. The primary contribution is an aerial-ground multi-agent control architecture that enables the safe and scalable deployment of coupled aerial and ground multi-robot operations. Safety across the aerial and ground layers is enabled through: (a) the design of a CBF-based controller for aerial and ground agents for enforcing collision avoidance constraints and (b) enabling the UAVs to safely land on their respective moving UGVs using a novel time-varying landing CBF. The scalability of the proposed aerial-ground coordination approach is enhanced through a hybrid centralized-distributed Edge architecture, where all the local agent controller nodes are distributed clients of a centralized Edge server node, which accumulates all the information to decide and communicate the relevant constraints to each of its clients.

The rest of the article is organized as follows: the model of the system and the control problem are formulated in Section \ref{sec:prob_form}, Section \ref{sec:cont_arch} presents the overall Control Architecture while the CBF filtering layer of the architecture is presented in Section \ref{sec:cbf_filt}, and the concluding remarks are presented in Section \ref{sec:conc}.

\section{Problem Formulation} \label{sec:prob_form}
In this section, the aerial (UAV) and ground (UGV) robots are modeled, the constraints necessary to enable safe coordination are identified, and the overall control problem is formulated.

\subsection{Modeling of the UAV}
\begin{figure}[!h]
	\centering
	\includegraphics[width=\linewidth]{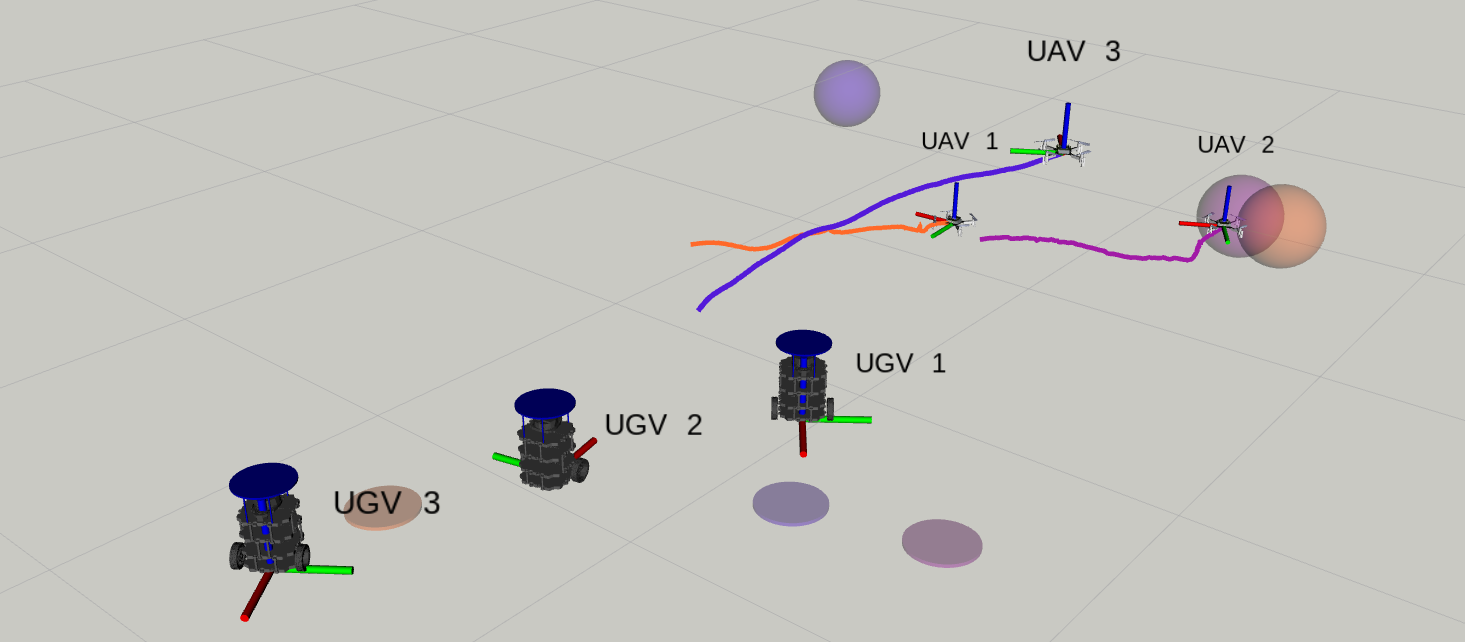}
	\caption{{A schematic of multi-robot task scenario.}}\label{fig:uav_model}
\end{figure}
The dynamics of a UAV with its body frame, denoted as $\mathbf{X_{ai} - Y_{ai} - Z_{ai}}$ is given by,
\begin{subequations} \label{eq:uav_model}
\begin{align}
    m_{i} {\ddot{p}_{ai}}(t) + m_{i} {G} &= f_{i}, \label{eq:p_tau} \\
    {J_{i}}(q_{i}(t)) {\ddot{q}_i}(t) + {C_i}({q_i, \dot{q}_i, t}){\dot{q}_i} (t) &= {\tau}_{{i}}, \label{eq:q_tau} 
\end{align}    
\end{subequations}
where $p_{ai}(t) \triangleq \begin{bmatrix}
    x_{ai}(t), y_{ai}(t), z_{ai}(t)
\end{bmatrix}^T \in \mathbb{R}^3$, $q_i(t) \triangleq \begin{bmatrix}
    \phi_i(t), \vartheta_i(t), \psi_i(t)
\end{bmatrix}^T$ represents the position and orientation of the UAV in the inertial frame ($\mathbf{X_W - Y_W - Z_W}$); $G \triangleq \begin{bmatrix}
    0, 0, -g
\end{bmatrix}^T \in \mathbb{R}^3$ is the gravity vector; $f_i  \in \mathbb{R}^3$ is the position control inputs in the inertial frame, (cf. Fig. \ref{fig:uav_model}); $m_i$ is the mass and $J_i, C_i \in \mathbb{R}^{3 \times 3}$ are the inertia and Coriolis matrices; $\tau_{i} \in \mathbb{R}^3$  is the attitude control input of the $i^{th}$ UAV for $i \in \lbrace 1, \cdots, N \rbrace $ and $N$ is the total number of UAVs.
\subsection{Modeling the Ground Robot}
Since UGVs in general are non-holonomic, we consider a differential-drive unicycle model for the UGVs as given below.
\begin{align}
    \begin{bmatrix}
        \dot{x}_{gi} \\ \dot{y}_{gi} \\  \dot{\theta}_{i} 
    \end{bmatrix} &= \begin{bmatrix}
        \cos(\theta_{i}) & 0 \\\sin(\theta_{i}) & 0  \\ 0 & 1
    \end{bmatrix} \begin{bmatrix}
        v_i \\ \omega_i
    \end{bmatrix}~~~ , ~~~  \begin{bmatrix}
        v_i \\ \omega_i
    \end{bmatrix} =  \begin{bmatrix}
        \frac{R_{1i} + R_{2i}}{2} \\ \frac{R_{1i} - R_{2i}}{2L}
    \end{bmatrix} \label{eq:ugv_model}
\end{align}
where $\begin{bmatrix}
    x_{gi} & y_{gi} 
\end{bmatrix}^\top$ form the position vector, $\theta_{i}$ represents the heading of the UGV in the inertial frame, $v_{i}, \omega_{i}$ represent the linear and angular velocities in the body frame, and $R_{1i}, R_{2i}$ represent the velocities of the right and left wheels respectively of the $i^{\text{th}}$ UGV and $L$ is the distance between the two wheels of the UGV. The body frame is represented by $\mathbf{X_{gi} - Y_{gi} - Z_{gi}}$.

\subsection{Constraint Identification}
In this article, we consider the scenario where there are $N$ UAV-UGV pairs that perform their predefined tasks. When a landing signal is provided, the UAVs return to their corresponding UGVs and land on them while the UGVs continue to perform their tasks. Therefore, the constraints can be broadly classified into interactions in the aerial layer (AA) between the UAVs, the interactions between the ground layer (GG) between the UGVs, and the aerial-ground interactions (AG) between the UAVs and the UGVs. The latter is further classified into the interaction of a UAV with its corresponding UGV (AGC) and the interactions of the UAV with other UGVs (AGO). It is intuitive that AA, GG and AGO are merely collision avoidance constraints between different agents for safely performing their high-level tasks. The AA constraints for the UAV-pair $(i, j)$ is represented with their relative displacements $r_{aij}$ as,
\begin{align}
    r_{aij}^2 \geq s_{a}^2, ~ r_{aij} = p_{ai} - p_{aj} \label{eq:aa_cons}
\end{align}
where $s_{a} > 0$ is the desired safety radius between two UAVs. Similarly, for AGO constraint, the relative displacement $r_{ij}$ between $i^{\text{th}}$ UAV and $j^{\text{th}}$ UGV must be outside a safety radius, $s_{ag} > 0$ given by,
\begin{align}
    r_{ij}^2 \geq s_{ag}^2, ~ r_{ij} = p_{ai} - p_{gj}. \label{eq:ago_cons}
\end{align}
Since ground robots are restricted to two dimensions, GG constraint becomes a circle in the $XY$-plane given by,
\begin{align}
    r_{gij}^2 \geq s_{g}^2, ~ r_{gij} = \rho_{gi} - \rho_{gj}, \label{eq:gg_cons}
\end{align}
where $s_{g} > 0$ is the safety radius, and $\rho_{gi}= \begin{bmatrix}
        x_{gi} & y_{gi}
    \end{bmatrix}^\top$ is the horizontal position of the UGV. Unlike these constraints, AGC cannot be simplified into a collision avoidance problem because while the constraint must restrict the $i^\text{th}$ UAV from colliding with the $i^\text{th}$ UGV, it must still allow the UAV to approach the UGV vertically and establish a contact. Hence, we extend our descending constraint from \cite{sankaranarayanan2024cbf} to a moving target scenario, which demands a time-varying constraint.
\begin{align}
    r_{zi}(t) &>=  \beta \alpha l_i(t)\exp(- \alpha l_i(t)) + \gamma, \label{eq:agc_constraint} 
\end{align}
where $(r_{xi}, r_{yi}, r_{zi})^\top \triangleq p_{ai} - p_{gi} \in \mathbb{R}^3$ define the relative displacement between the $i^\text{th}$ UAV-UGV pair, $l_i = r_{xi}^2(t) + r_{yi}^2(t)$ is their corresponding squared horizontal distance, $\alpha, \beta$ are scaling factors in the horizontal and vertical directions respectively and $\gamma$ is an offset to avoid the turbulence from the landing platform. In this setup, Eq. \eqref{eq:ago_cons} distances other UAVs from the $i^\text{th}$ UGV clearing out the space around the $i^\text{th}$ UGV for a conflict-free landing. Additionally, we apply box constraints on all the agents to replicate an indoor environment with walls and roof as given below,
\begin{subequations}
\begin{align}
    \underline{x} &< x_{ai} < \overline{x}, ~ \underline{y} < y_{ai} < \overline{y}, ~\underline{z} < z_{ai} < \overline{z}, \label{eq:box_uav} \\
    \underline{x} &< x_{gi} < \overline{x}, ~ \underline{y} < y_{gi} < \overline{y}, \label{eq:box_ugv}
\end{align} \label{eq:box_cons}   
\end{subequations}
where $\underline{x}, \overline{x}, \underline{y}, \overline{y}, \underline{z}, \overline{z}$ form the maximum and minimum bounds for the states. Further, we highlight the necessary assumptions for formulating the control problem.
\begin{assum} [Minimum distance between UGVs] \label{as:min_dist}
    The safety radii magnitudes must be such that, $s_{g} > s_{ag} > s_{a}$.
\end{assum}
\begin{assum} [Velocity of UGVs] \label{as:vel}
    The linear and angular velocities of the UGVs are smooth and bounded, such that $|v_{i}| < \overline{v}, |\omega_{i}| < \overline{\omega}$. Further, the UGVs move on a flat surface that $\dot{z}_{gi} = 0 ~ \forall i \in \lbrace 1,2, ..., N \rbrace$.
\end{assum}
\begin{remark}
    Assumption \ref{as:min_dist} is standard, since the landing platforms are sufficiently larger than the cross-sectional area of the UAVs. In its absence, the AA and AGO constraints will restrain the UAVs from landing on the UGVs. Further, the upper bound on the velocities in Assumption \ref{as:vel} translates to $|\dot{x}_{gi}|, |\dot{y}_{gi}| < \overline{v}$, which is essential for the UAV to track and land on its UGV.
\end{remark}

While latency and momentary loss of communication make an offboard control scheme unreliable, the onboard controllers are limited by computing power. Since the number of constraints per UAV is $2N + 4$ for $N$ UAV-UGV pairs ($N+3$ for UGVs), it is optimal to consider only the proximal interactions (within a safety distance) for AA, GG and AGO. However, in fully connected distributed systems, deciding the number of constraints also requires the knowledge of every agent, which increases the computational load and number of connections between the agents. Moreover, the interaction constraints AA, GG, AGO, and AGC in \eqref{eq:aa_cons} - \eqref{eq:agc_constraint} are non-convex in nature. Given these challenges, let us formulate the control problem.

\noindent \textbf{Problem Formulation:} Under the assumptions \ref{as:min_dist} $\&$ \ref{as:vel}, design a control architecture for the coordinated operation of multi UAV-UGV pairs, whose dynamics is presented in Eq. \eqref{eq:uav_model} - \eqref{eq:ugv_model}, while enforcing the constraints associated with collision avoidance and safe landing of UAVs on their respective mobile UGVs formulated in Eq. \eqref{eq:aa_cons} - \eqref{eq:box_cons} .

\section{Edge-enabled Control Architecture} \label{sec:cont_arch}
Following the control challenge, we introduce the proposed control architecture, which suitably accommodates the constraints and minimizes the onboard computations and number of connections between the agents.

\subsection{Edge Architecture} \label{subsec:edge_imp}
\begin{figure}[!h]
\centering
    \includegraphics[width=0.8\linewidth]
    {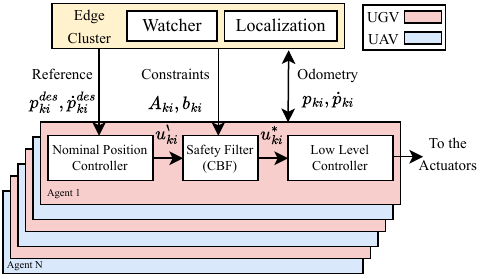}
    \caption{Edge architecture. Each robot has its own distributed control unit, and receives CBF constraints and its position estimation from the edge cluster.}
    \label{fig:star_topology}
\end{figure}
In the considered scenario, edge computing provides the benefit of distributed onboard control and a centralized decision processing unit \cite{robotics11040080}. Here, the UAVs and UGVs function as edge devices (clients) of the system, while an edge cluster (server) provides the computational capacity \cite{seisa2022edge}. The control architecture consists of three main segments: the CBF-enabled control unit, the data acquisition and localization unit, and a remote monitoring and decision-making unit (referred as Watcher). The control units are distributed to run onboard on the edge clients. Since computational power is a trade-off on the onboard computers, we use first-order CBFs to impose the safety constraints. The design of the control unit is further discussed in Subsection \ref{subsec:control_unit}. 

We deploy $N$ UAV-UGV pairs ($2N$ agents) and establish an edge cluster to offload the localization method of all agents, along with the Watcher, as depicted in Fig.~\ref{fig:star_topology}. We enable communication between the agents through Wi-Fi connectivity. The localization unit either uses an external positioning system (as in our experimental setup) or performs a global SLAM using the agents' onboard sensor information to estimate the pose of each agent. The Watcher node accesses the localization unit and identifies the proximal agents to each of the robots and generates its constraint matrix. As such, each agent's control unit requires only its global localization and setpoints for the nominal controller, and the constraint matrices to enforce safety. Since no further information is required for safe interaction among agents, the proposed topology supports a simplified connectivity structure, thereby facilitating streamlined communication, enabling scalability, and reducing throughput.

The scalability of the proposed architecture is driven by two main factors, thanks to the hybrid nature of edge computing. (a) Number of communication links: The proposed star topology requires links equal to the number of agents, i.e., $2N$ (UAV and UGV), since agents exchange information only with the Watcher. In contrast, in fully connected topologies, the number of links is equal to $2N\times(2N-1)$. In addition, systems that enable communication only among neighboring agents need to reestablish their communications constantly and dynamically. (b) Computational capacity: Distributed computational processing allows for effective management of computational resources depending on the number of proximal agents. 

\subsection{Distributed Control Units} \label{subsec:control_unit}
This section presents the distributed control unit for UAVs and UGVs. The control unit is divided into three layers: a nominal position controller, a CBF-based safety filter, and a low-level controller. The position controller produces nominal velocity inputs to reach the setpoints $(p_{ai}^{des} \in \mathbb{R}^3, \rho_{gi}^{des} \in \mathbb{R}^2)$ given by, $u_{ai}^{'} = -K_{ai}(p_{ai} - p_{ai}^{des}) + \dot{p}_{ai}^{des},~ u_{gi}^{'} = -K_{gi}(\rho_{gi} - \rho_{gi}^{des}) + \dot{\rho}_{gi}^{des}$, where $K_{ai}, K_{gi}$ are positive definite gain matrices. This input is filtered using first-order CBF constraints to obtain the filtered velocity input $u_{ai}^* \in \mathbb{R}^3, u_{gi}^*\in \mathbb{R}^2$ as will be explained in Section \ref{sec:cbf_filt}. The UAV's filtered velocity inputs are tracked by a low-level controller (cf. \cite{sankaranarayanan2024cbf} Section III).

Since the CBF filter uses first-order constraints, it uses the first-order kinematics of the robots during optimization. The UAV's first-order kinematics simplifies to $\dot{p}_{ai} = u_{ai}(t)$. However, since UGV is non-holonomic, we use the Near-identity diffeomorphism (NID) presented in \cite{wilson2020robotarium, olfati2002near} to get its single integrator approximation given by,
\begin{align}
    \begin{bmatrix}
        x_{oi} \\ y_{oi}    
    \end{bmatrix} & = \begin{bmatrix}
        x_{gi} \\ y_{gi}   
    \end{bmatrix} + o_i \begin{bmatrix}
        \cos(\theta_i) \\ \sin(\theta_i)
    \end{bmatrix},    \label{eq:NID} \\
    \implies \begin{bmatrix}
        v_{i} \\ \omega_{i}
    \end{bmatrix} &=  \begin{bmatrix}
        \cos(\theta_{i}) & \sin(\theta_{i}) \\
        -\frac{\sin(\theta_{i})}{o_i} & \frac{\cos(\theta_{i})}{o_i} 
    \end{bmatrix} \begin{bmatrix}
             \dot{x}_{oi} \\ \dot{y}_{oi}    
        \end{bmatrix} \label{eq:NID_inv}
\end{align}
where $o_i > 0 \in \mathbb{R}^+$. Geometrically, $\rho_{oi} = \begin{bmatrix}
    x_{oi} & y_{oi}
\end{bmatrix}^\top$ represents a 2D point (offset) at a distance of $o_i$ from the $i^\text{th}$ UGV's origin along the $\mathbf{X_{gi}}$ axis. By choosing a sufficiently small $o_i$, the same constraints could be used in the offset space. It is to be remarked that converting the dynamics using NID retains the same nominal control input, which is passed on to the CBF layer. The filtered velocity input $u_{gi}^* \in \mathbb{R}^2$ obtained from the CBF layer is reconverted to the UGV's body frame using Eq. \eqref{eq:NID_inv}, which is used for actuating the wheel motors based on the relationship in \eqref{eq:ugv_model}.

\section{CBF Implementation} \label{sec:cbf_filt}
Now that the overall control architecture is described, this section presents the CBF filter implementation inside a distributed control node and its constraint matrices. As discussed in the earlier sections, though the safety constraints are imposed by the distributed control node through quadratic programming, the constraint matrices for every distributed control unit are updated by the Watcher node using ROS (Robot Operating System) messages. Since every distributed control unit handles the control inputs of only the specific robot, the constraints that use the position of other agents become time-varying. So, let us formally define the time-varying control barrier function before formulating the CBFs, and subsequently forming the quadratic program.

Let the time-varying safe set for a given affine dynamical system, $\dot{x}(t) = f(x(t)) + g(x(t))u(t)$ be defined by the state space region, $\mathcal{S}(t) \subset \mathcal{X}$, where $x \in \mathcal{X} \subset \mathbb{R}^n, u \in \mathbb{U} \subset \mathbb{R}^m, f, g$ are Lipchitz continuous. Now, a continuously differentiable function $h(x,t): \mathcal{D}(t) \subset \mathcal{X} \xrightarrow{} \mathbb{R}$, such that $\mathcal{S}(t):\lbrace x \in \mathcal{X} | h(x,t) \geq 0 \rbrace$ ($\mathcal{S}(t)$ is a zero-level super set of $h(x,t) \forall t$), renders the set $\mathcal{S}(t)$ safe if the control input $u$ ensures positive invariance of the set $\mathcal{S}(t) ~ \forall t$, i.e.,  $x(t_0) \in \mathcal{S}(t_0)$ implies $x(t) \in \mathcal{S}(t) ~ \forall t \geq t_0$. Moreover, if the safe set, $\mathcal{S}(t)$ is rendered asymptotically stable when $x(t)$ is initialized outside the unsafe set ($x(0) \in \mathcal{D}(0) \setminus \mathcal{S}(0)$), a measure of robustness can be incorporated into the notion of safety. The condition validating the continuous function, $h(x,t)$ as a control barrier function, is presented in the following definition.
\begin{definition}[Time-varying CBF] \label{Def:TV-CBF}
    A candidate function $h(x,t)$ is a valid control barrier function for an affine dynamical system, $\dot{x} = f(x,t) + g(x,t)u$ if there exists a locally Lipschitz continuous class-$\mathcal{K}_\infty$ function $\xi$, such that $\forall (x,t) \in \mathcal{D}(t) \times [0, \infty]$
    \begin{align}
        \sup_{u \in \mathcal{U}} \frac{\partial h(x,t)^T}{\partial x} \left (f(x) + g(x)u  \right ) + \frac{\partial h(x,t)}{\partial t} \geq -\xi(h(x,t)) \label{eq:cbf_def}
    \end{align}
\end{definition}
\begin{remark}
The condition in \eqref{eq:cbf_def} captures the forward invariance of $\mathcal{S}(t) ~(\dot{h}(x,t) \geq 0 \text{ on } \partial \mathcal{B}(t))$, and the asymptotic stability of $\mathcal{S}(t) ~ (\dot{h}(x,t) \geq 0 \text{ in } \mathcal{D}(t)\setminus \mathcal{S}(t))$.
\end{remark}

Now, the CBF design for the safety constraints are discussed in Subsections \ref{subsec:sph_cbf} - \ref{subsec:box_cbf}. Subsequently, the quadratic programming is formed with appropriate constraints in Subsection \ref{subsec:qp}.

\subsection{Spherical CBF} \label{subsec:sph_cbf}
The inequalities in \eqref{eq:aa_cons} - \eqref{eq:ago_cons} form spherical and \eqref{eq:gg_cons} form circular constraints for the robots. Therefore, we derive the following candidate functions. 
\begin{align}
    h_{aij} = r_{aij}^2 - s_a^2, ~ h_{ggij} = r_{oij}^2 - s_{g}^2, ~ h_{agij} = r_{ij}^2 - s_{ag}^2, \label{eq:sph_cbf}
\end{align}
where $r_{oij} = (\rho_{oi} - \rho_{oj})$ is the relative distance between the offsetted positions of the UGVs. It is to be remarked that since these constraints depend on the time-varying positions of other robots, which are external to the particular robot's system, the CBFs are time-varying.
Through Definition \ref{Def:TV-CBF}, it can be shown that the candidate CBFs \eqref{eq:sph_cbf} are valid time-varying CBFs for the UAV's and UGV's respective admissible control sets of $\mathcal{U}_a = [-\overline{v}, \overline{v}] \times [-\overline{v}, \overline{v}] \times [-\overline{v}, \overline{v}]$ and $\mathcal{U}_g = [-v_g, v_g] \times [-v_g, v_g]$, where $0 < v_g < \overline{v}$. Provided the UAVs and UGVs initiate with a distance between them larger than their respective safety radii, they do not get closer than their respective safety radii throughout the time, thus avoiding possible collisions, due to the virtue of forward invariance of the CBFs.

\subsection{Landing CBF} \label{subsec:land_cbf}
For the AGC constraint, a candidate function is formed as,
\begin{align}
    h_{lii} = r_{zi} - \beta\,\alpha\, l_i \,\exp(- \alpha l_i) - \gamma \label{eq:lcbf}
\end{align}
Unlike \cite{sankaranarayanan2024cbf}, $h_{lii}$ is time-varying as it depends on an UGV's time-varying position. Further, their gradients are given by,
\begin{align}
    \frac{\partial h_{lii}}{\partial p_{ai}} &= \begin{bmatrix}
        k\,r_{xi} & k\,r_{yi} & 1
    \end{bmatrix}^\top, ~ \frac{\partial h_{lii}}{\partial t} = -k\,(r_{xi}\dot{x}_{gi} + r_{yi}\dot{y}_{gi})
\end{align}
where $k = 2\beta \alpha (\alpha l_i - 1) \exp(-\alpha l_i)$. Therefore, the candidate function in \eqref{eq:lcbf} can be proven to be a valid time-varying CBF using Definition \ref{Def:TV-CBF} for the UAV's admissible control set of $\mathcal{U}_a = [-\overline{v}, \overline{v}] \times [-\overline{v}, \overline{v}] \times [-\overline{v}, \overline{v}]$. Since the CBF possesses the forward invariance property, when the UAVs are initiated in the safe region defined by the zero super level set $\mathcal{S}_i = \lbrace p_{ai} \in \mathbb{R}^3 | h_{lii, t} \geq 0 \rbrace$, they remain in the safe region during the landing process, which forces them to vertically descend through a funnel-like region above the landing platform.

\subsection{Box CBF} \label{subsec:box_cbf}
The box inequalities in \eqref{eq:box_cons} form the following functions.
\begin{subequations}
    \begin{align}
        b_{a1i} &= \overline{x} - x_{ai}, ~ b_{a2i} = x_{ai} - \underline{x}, ~ b_{a3i} = \overline{y} - y_{ai}, \\ b_{a4i} &= y_{ai} - \underline{y}, ~ b_{a5i} = \overline{z} - z_{ai},  
        ~ b_{g1i} = \overline{x} - x_{oi}, \\ b_{g2i} &= x_{oi} - \underline{x}, ~ b_{g3i} = \overline{y} - y_{oi}, ~ b_{g4i} = y_{oi} - \underline{y}.
    \end{align} \label{eq:box_cbf}
\end{subequations}
A separate CBF for the lower bound of the UAV's altitude is redundant, as the AGC constraint in \eqref{eq:lcbf} maintains the UAV above the landing platform \cite{sankaranarayanan2024cbf}. Unlike other constraints, the box constraints are time-invariant. 
Using similar arguments in the previous subsections, the functions in \eqref{eq:box_cbf} are proven to be valid control barrier functions. So, the UAVs and UGVs remain inside the box if they are initiated within it.

\subsection{CBF Safety Filter for $i^\text{th}$ Agent} \label{subsec:qp}
The safety filter running on the distributed control units takes in the nominal input $u_{ki}^{'}\left(p_{ki}(t), t\right) ~ \forall k \in \lbrace d, g \rbrace$, and generates a filtered input $u_{ki}^*$, which satisfies the safety constraints. This action is performed by a quadratic program:
\begin{align}
    u_{ki}^*\left(p_{ki}(t), t \right) &= \underset{u_{ki} \in \mathcal{U}_k}{\text{argmin}} || u_{ki} - u_{ki}^{'}\left(p_{ki}(t), t\right) || \nonumber \\
    s.t. &: \mathbf{A}_{ki} u_{ki} \geq - \mathbf{b}_{ki}. \label{eq:qp}
\end{align}
The constraint in \eqref{eq:qp} is derived using \eqref{eq:cbf_def} and the first-order approximations of the robots, where $f(x) = 0, g(x) = \mathbf{I}, \mathbf{A} = \frac{\partial h}{\partial x}, \mathbf{b} = \xi(h,t) + \frac{\partial h}{\partial t}$. As mentioned in the control architecture, the matrices $\mathbf{A}_{ki}, \mathbf{b}_{ki} ~\forall k \in \{ a,g \}$ are generated for each control unit by the Watcher node. $\mathbf{A}_{ki}, \mathbf{b}_{ki}$ are initialized with zeros of size $(C \times 3), (C)$ respectively. Then the boundary constraints are updated to $\mathbf{A}_{ki}, \mathbf{b}_{ki}$, followed by AG or GG constraints. If it is an UAV, AGC and AA constraints are added.


The partial derivatives of the CBFs in \eqref{eq:sph_cbf} \& \eqref{eq:box_cbf} are not explicitly mentioned, because they are trivial.

\section{Conclusion} \label{sec:conc}
In this work, we have discussed the preliminary control architecture for safe and scalable coordination of heterogeneous robotic systems. We have identified the necessary constraints and formulated CBF-based controller for collision avoidance, safe landing, and task space restriction. We have proposed an Edge-based hybrid architecture to reduce the computational load in the onboard controller. The validation of the proposed controller and theoretical evaluation will be included in the extended version.

\bibliographystyle{IEEEtran}
\bibliography{ref}
\end{document}